\documentclass[11pt,letterpaper]{article}

\usepackage{emnlp2017}
\usepackage{times}
\usepackage{latexsym}
\usepackage{graphicx}
\usepackage{linguex}
\usepackage[plain]{fancyref}
\usepackage{textcomp}

\emnlpfinalcopy

\def\emnlppaperid{57}

\title{The Helsinki Neural Machine Translation System\Thanks{The software is
available from \url{https://github.com/robertostling/hnmt} under the GNU
General Public License version 3.}}

 \author{Robert {\"O}stling \\ Department of Linguistics \\ Stockholm University
         \And 
         Yves Scherrer \and J{\"o}rg Tiedemann \\ Department of Modern
     Languages \\ University of Helsinki \AND
        Gongbo Tang \\ Department of Linguistics and Philology \\ Uppsala
    University \And
    Tommi Nieminen \\ Department of Modern 
     Languages \\ University of Helsinki}

\date{}

\begin{document}

\maketitle

\begin{abstract}
We introduce the Helsinki Neural Machine Translation system (HNMT) and how it
is applied in the news translation task at WMT 2017, where it ranked first in
both the human and automatic evaluations for English--Finnish.  We discuss the
success of English--Finnish translations and the overall advantage of NMT over
a strong SMT baseline. We also discuss our submissions for English--Latvian,
English--Chinese and Chinese--English.
\end{abstract}

\section{Introduction}

The Helsinki Neural Machine Translation system (HNMT) is a full-featured
system for neural machine translation, with a particular focus on
morphologically rich languages. We participated in the WMT 2017 shared task on
news translation, obtaining the highest BLEU score for English--Finnish
translation, while also performing well on English--Latvian and acceptably on
English--Chinese and Chinese--English.

In addition to our participation in the shared task, this paper also details
some of the other methods we have implemented and evaluated with HNMT, many of
which yielded negative results and were subsequently not used in our
submissions for the shared task.

\subsection{HNMT}

HNMT is based on the
attentional encoder--decoder model due to \citet{Bahdanau2014nmt}. This is a
rather minimalistic framework for NMT, and many extensions have been proposed.
Of particular interest are those that allow proper and efficient handling of
morphologically rich languages, such as Finnish. We combine two such
approaches: the hybrid character/word model of \citet{Luong2016achievingopen},
which is used for the source language encoder, and the byte-pair encoding
(BPE) technique of \citet{Sennrich2016subword}, which is used for the target
language decoder and has been successfully used for Finnish previously
\citep{sanchez2016abu}. As BPE can be added as a simple pre- and post-processing
step, it does not affect the structure of the translation model. This means
that our system can be used with character, word and BPE
level generation on the target side.
The structure of the network, thus, consists of three Long Short-Term Memory (LSTM)
 \citep{Hochreiter1997lstm} layers:
\begin{enumerate}
    \item A character-level encoder that transforms out-of-vocabulary
        source
        tokens into the same vector space as the source word embeddings.
    \item A token-level bidirectional encoder that transforms a
        sequence of
        source word embeddings (or outputs from network (1) in case
        of OOV items) into an encoded sequence of the same length.
    \item A character-, token- or BPE-level decoder that works as language
        model conditioned (via an attention mechanism) on the encoded source
        sequence from (2).
\end{enumerate}
HNMT is implemented using Theano \citep{Theano2016}, which supports
efficient training with a GPU.  For optimization we use minibatch stochastic
gradient descent with Adam \citep{Kingma2014adam} for learning rate
adaptation.

\section{Tricks from the NMT arsenal}

We have implemented and evaluated a number of proposed extensions to the basic
attentional encoder--decoder model. Basic experiments were carried out on
English--Finnish data, unless specified otherwise.

\subsection{Layer normalization}

Layer normalization \citep{Ba2016layernormalization} has been proposed as a
technique for speeding up training of recurrent network models. We have
implemented it into HNMT as the modified LSTM described by
equations (20)--(22) in \citet{Ba2016layernormalization}. However, as
preliminary experiments did not indicate any consistent effect of using layer
normalization we did not include it in our evaluation.

\subsection{Variational dropout}

\citet{Gal2016theoretically} proposed a method for regularization of recurrent
neural networks. This has also been implemented in HNMT, but preliminary
experiments on Finnish did not indicate any improvement over the baseline
system. While \citet{Sennrich2016uedinwmt} reported large improvements for
the Romanian news translation task at WMT 2016, the amount of training data is
lower than what is available for Finnish, which should explain some of the
difference. They also apply dropout on the word level, whereas the HNMT
application currently only drops recurrent states.

\subsection{Context gates}

Context gates \citep{Tu2017contextgates} introduce an explicit model for
selecting to which extent the target sentence generation should focus on the
source sentence or the target context, giving the network a chance to tune the
balance between adequacy and fluency. While we obtained better cross-entropy
on the development set, particularly early during training, the BLEU and chrF3
evaluations on development data made us decide against the slower context
gates in the final run.


\subsection{Coverage decoder}

\citet{Wu2016gnmt} present an empirically determined method for using
the attention vectors produced during decoding in the search algorithm, to
bias the decoder towards translations with reasonable length and good
coverage of the source sentence.\footnote{The HNMT implementation of this
was contributed by Stig-Arne Gr{\"o}nroos.}
We performed a grid search of the parameter space for the length, coverage and
overtranslation penalties, but did not find any that resulted in higher BLEU
scores on the development set than the decoder without penalties.

\subsection{Forward-Backward reranking}

It is trivial to train a translation model to generate translations either
from the beginning or the end of the target sentence. HNMT supports selecting
translation direction, and combined with its n-best list and reranking
features it is simple to generate candidate translations in both directions and to combine them based on their scores.
This led to some minor improvements in our English--Finnish translations.

\subsection{Ensembling}

HNMT supports two general modes of ensembling, as well as their combination:
\begin{itemize}
    \item Proper ensembling where
        $p(w) = \frac{1}{M} \sum_{m=1}^M p_m(w)$ is
        used to predict target symbol $w$, given predictions $p_m(w)$ for
        each model $m$ in the ensemble.
    \item Parameter averaging where the model's parameter vector $\theta$ is
        computed as $\frac{1}{N} \sum_{m=1}^M \theta_m$ for each model $m$.
        This only works if the different $\theta_m$ are relatively similar,
        typically because they were saved at different points during the same
        training process.
\end{itemize}
The overhead for proper ensembling is linear in the number of ensembled
systems, both for training (assuming one is building an ensemble of separately
trained models) and inference, while parameter averaging is
essentially free. HNMT allows proper ensembling of groups of models where the
parameters are averaged within each group. This flexible structure allows a
number of setups, which are explored further in \Fref{sec:enfi-nmt}.

\section{English--Finnish}
\label{sec:enfi}

In our experiments, we used all English--Finnish parallel data sets provided
by WMT except the Wiki headlines, which is a small and rather noisy data set
that did not contribute anything in our experiments from last year. We also
added substantial amounts of backtranslated data that has been shown to help
especially in neural machine translation \cite{Sennrich2016backtranslation}
but also in statistical MT \cite{tiedemann-EtAl:2016:WMT}.
Table~\ref{tab:backtrans-fien} lists some basic statistics of the
backtranslated data sets we created out of WMT's monolingual Finnish news data
from 2014 and from 2016. We applied our best constrained phrase-based SMT
model for Finnish to English from last year \cite{tiedemann-EtAl:2016:WMT}
that uses a factored model with multiple translation paths, morphological tags
and pseudo-tokens for case-markers that correspond to English prepositions
\cite{Tiedemann_etal:15}. The system scored 20.5\% lowercased BLEU on the
newstest 2016 data, which was the second-best system for the task in 2016.

\begin{table}[ht]
\small \centering
\begin{tabular}{|l|ccc|}
\hline
                  & sentences &    Finnish    & English\\
\hline
news2014  & 1,378,833 & 17,117,137 & 23,818,547\\
news2016  & 4,144,406 & 55,637,304 & 76,161,439\\
\hline
\end{tabular}
\caption{Backtranslated Finnish news data.}
\label{tab:backtrans-fien}
\end{table}

\subsection{Preprocessing and postprocessing}

We trained our models on tokenized and truecased data, except for the character-level models which were trained on raw untokenized data. For the former, we applied Moses tools for Unicode/punctuation normalisation, tokenization and truecasing using a model trained on the parallel training data.

We tested three different types of word segmentation: basic word-based segmentation, supervised morphological segmentation using OMorFi \cite{pirinen2015development} and byte-pair encoding (BPE) \cite{Sennrich2016subword}. For the latter, we opted for a fine-grained segmentation that results in a small vocabulary of 20,000 tokens when trained on the parallel data, expecting BPE to handle various cases of compound splitting and morphological segmentation. We always used the same BPE-based segmentation and did not try to optimize the BPE parameters in any way. 

During development, we observed that the English development files contained a lot of verb form contractions (of the type \textit{wouldn't}), but that such contracted forms appear rarely in the training data. Therefore, we also added a preprocessing routine to transform the contracted forms to their uncontracted equivalents.

Finally, we found that our tokenizer/detokenizer pipeline for Finnish did not handle the hyphen/dash distinction correctly. In Finnish, the `-' sign can be used with spaces on both sides, without spaces, with a space only on the left, and with a space only on the right, as in the following examples:

\ex.
\a. Draamaa Riossa - suomalaisnostaja py\"{o}rtyi\dots \\
`Drama in Rio - Finnish lifter fainted\dots'
\b. Kempinski-hotelli \\
`Kempinski[-]hotel'
\c. kissa ja hiiri -leikki\"{a}\\
`cat and mouse [-]game'
\d. \"{o}ljy- ja kaasutoiminnot\\
`oil[-] and gas functions'

The tokenizer always introduces spaces on both sides, which means that the detokenizer is then unable to retrieve the original configuration. 
In order to remedy this problem, we applied a postprocessing step to the translated data. After detokenizing the output, for every hyphen sign, the four tokenization variants were generated and scored by the hybrid-to-character system; we then chose the tokenization variant with the highest score.

\subsection{NMT Models}
\label{sec:enfi-nmt}

In preliminary experiments, we focused on different segmentation strategies
for the source and target sides as well as on different proportions of
backtranslations and parallel data. The models were evaluated on \textit{newsdev2015}
using lowercased BLEU and chrF3.\footnote{The HNMT-internal BLEU computation is based on \url{https://github.com/vikasnar/Bleu} and on the NLTK tokenizer. The reported results are thus not directly comparable with official WMT results.} Table~\ref{tab:NMT-enfi-dev} shows some results.

\begin{table*}
\small \centering
\begin{tabular}{|ll|cccc|cccc|}
\hline
&  & \multicolumn{4}{c|}{BLEU} & \multicolumn{4}{c|}{chrF3} \\
Encoder & Decoder & None & Only & Balanced & All & None & Only & Balanced & All \\
\hline
BPE & BPE		& 11.9 & \bf 14.4 & \bf 15.7 & \bf 15.5 & 43.7 & 47.2 & 48.3 & 48.5 \\
BPE & Char		&  9.2 & 13.0 & 13.7 & 14.0 & 41.0 & \bf 47.8 & \bf 48.4 & 48.6 \\	
Hybrid & BPE	& \bf 12.2 & 13.8 & 15.4 & 15.3 & 43.4 & 47.0 & 48.1 & 49.0 \\
Hybrid & Char	& 11.6  & 13.1 & 14.1 & 14.2 & \bf 46.3 & 47.2 & 48.2 & 49.0 \\	
Hybrid & OMorFi	& ---  & ---  & ---  & 14.3 & ---  & ---  & ---  & \bf 49.2 \\
\hline
\end{tabular}
\caption{Development results with different segmentation strategies for the source language encoder and the target language decoder and different proportions of backtranslated and parallel data (None = 2.5M sentences of parallel data + 0 sentences of backtranslated data; Only = 0 + 5.5M; Balanced = 2.5M + 2.5M; All = 2.5M + 5.5M).}
\label{tab:NMT-enfi-dev}
\end{table*}

In these experiments, we found BPE to be useful on the target side, but not so much on
the source side. Character-level decoders are favoured by character-level evaluation scores such as chrF3, whereas BLEU favours decoders using larger units such as BPE. The best results were obtained with a combination of backtranslated and parallel data; using all backtranslations was slightly better than restricting the amount of backtranslations to match the size of the parallel data. The model based on supervised morphological segmentation
followed by BPE encoding (OMorFi) yielded promising chrF3 results, but lagged behind in terms of BLEU. Further investigation is needed on the benefits and shortcomings of combining these segmentation approaches.

The model based on a hybrid encoder and a BPE decoder did not yield the best results in this preliminary evaluation, but showed the most robust performance across different evaluation types, training configurations and evaluation data (in particular, it outperformed other models on the \textit{newstest2015} set). Therefore, four of our five submissions use that configuration. For comparison, we also submitted a system based on a character-level decoder. 

\begin{table}
\small \centering
\begin{tabular}{|rr|rrc|}
\hline
BLEU & chrF3 & M & SP/M & AVG \\
\hline
12.8 & 48.8 & 1 & 1 & N/A \\
13.6 & 49.7 & 1 & 4 & $+$ \\
13.8 & 49.8 & 1 & 4 & $-$ \\
14.1 & 50.0 & 3 & 1 & N/A \\
14.4 & 50.2 & 3 & 4 & $+$ \\
14.6 & 50.4 & 3 & 4 & $-$ \\
\hline
\end{tabular}
\caption{Development results with different ensembling setups. Each
configuration consists of M models, with SP/M savepoints per model, where the
savepoints may be averaged (+AVG) or included as equal members in the ensemble
(-AVG).}
\label{tab:NMT-enfi-dev-ensemble}
\end{table}

We also investigated the effect of different ensembling combinations, and the
result can be found in \Fref{tab:NMT-enfi-dev-ensemble}. In general,
proper ensembling is better than savepoint averaging, but savepoint averaging
is better than nothing. Further experiments revealed that the difference
between an \emph{ensemble of averaged savepoints from independent models}
setup (second row from the bottom) and an \emph{ensemble of several savepoints
each from independent models} (bottom row) is not consistent, so we use the
former (faster) variant for our official submissions.

The submitted character-decoder system uses 256 dimensions for word
embeddings, 64 for character embeddings, 512 encoder state dimensions, 1024
decoder state dimensions, and 256 attention dimensions.
We train four independent models for 72h each, and the savepoint with the best
heldout chrF3 score is used (in practice we do not observe any significant
overfitting, so this amounts to using nearly 72h of training for all models).
Training data are the unprocessed versions of all parallel and
backtranslated data. For decoding, we used proper ensembling of the four
models, and averaging of the four last savepoints of each model (states were
saved after each 5000 training batches).

The submitted BPE-decoder systems use the same model size as the
character-decoder system.
Again, we train four independent models for 72h each, using the preprocessed
and BPE-encoded data, with hyphen retokenization applied as a postprocessing
step. We provide two contrastive systems: one without input normalization,
which shows a decrease of 0.3 BLEU, and one without hyphen retokenization,
which shows a decrease of 0.9 BLEU (see Table~\ref{tab:NMT-enfi}).

We also propose an extended system that is based on the four models above and four additional backwards models (i.e., trained right-to-left). At test time, we generate a 10-best list of forward translations and a similar one of backward translations. We choose the best translation that occurred in both lists, or if the lists are disjoint, the translation with the highest likelihood according to the model (forward or backward) that generated it. This reranking only provided +0.1 BLEU; 48\% of translations were chosen from the forward model, 22\% from the backward model, and 30\% occurred in both lists. This system has been ranked first in the automatic and manual evaluations.

\begin{table}
\small \centering
\begin{tabular}{|lcccc|}
\hline
Decoder & IN & HR & Direction & BLEU \\
\hline
Char & $+$ & N/A & fw & 19.1 \\
BPE & $+$ & $+$ & fw & 20.6 \\
BPE & $-$ & $+$ & fw & 20.3 \\
BPE & $+$ & $-$ & fw & 19.7 \\
BPE & $+$ & $+$ & fw+bw & \bf 20.7 \\
\hline
\end{tabular}
\caption{Submitted HNMT systems with official results. They vary with respect
to decoder type, input normalization (IN), hyphen retokenization (HR),
direction (forward or backward). The best result was submitted for manual
evaluation, where it ranked \#1 (tied with one unconstrained system).}
\label{tab:NMT-enfi}
\end{table}

\subsection{SMT Baselines}

Besides the neural MT models, we also trained various phrase-based SMT models to contrast our results with another popular paradigm. In particular, we were interested to see the effect of BPE segmentation and backtranslation on statistical MT. Both techniques are popular in neural MT but their impact on statistical MT has not been evaluated properly before. Therefore, we started a systematic comparison of different setups including various types of segmentations and data collections. All systems are based on Moses \cite{Koehn_etal:07} and we use standard configurations for training non-factored phrase-based SMT models using KenLM for language modeling \cite{Heafield:11} and BLEU-based MERT for tuning. The only difference to the standard pipeline is the use of efmaral \cite{Ostling2016efmaral}, an efficient implementation of fertility-based IBM word alignment models with a Bayesian extension and Gibbs sampling.\footnote{Software available from \url{https://github.com/robertostling/efmaral}.}
Table~\ref{tab:SMT-enfi} summarizes the results of our SMT experiments during development.

\begin{table}[ht]
{\small
\begin{tabular}{|l|rrrcc|}
\hline
{\em newsdev15}                   & \multicolumn{3}{c}{segmentation} &  \multicolumn{2}{c|}{LM} \\
data            &   src    &    trg    &   tuning   & news & +CC \\
\hline
WMT           &    word      &    word     &   word      & {\bf 12.51}  & {\bf 13.74}\\
WMT           &    word      &    BPE       &   word      & 12.16  & -- \\
WMT           &    word      &    morf  &   word      & 11.58  & -- \\
WMT           &    BPE        &    BPE       &   BPE        & 11.91  & --     \\
WMT           &    BPE        &    BPE       &   word      & 12.24  & 12.95 \\
\hline
back           &    word      &   word      &  word       & 12.69  & {\bf 13.69} \\
back           &    BPE        &   BPE        &  BPE         & 12.73  & -- \\
back           &    BPE        &   BPE        &  word       & {\bf 12.92}  & 13.50 \\  
\hline
WMT+back &    word      &   word      &  word       & --       & {\bf 14.62} \\
WMT+back &    BPE        &   BPE        &  BPE         & 12.94  & -- \\
WMT+back &    BPE        &   BPE        &  word       & 13.40  & 14.44 \\
+osm & word &  word     &  word       & --        & 14.04 \\
+osm & BPE   &  BPE       &  word       & 12.85   & 14.58 \\
\hline
opus          &     word       &   word      & word     & 14.05  & 15.54 \\
opus          &     BPE         &   BPE        & word     & 14.45  & 15.63 \\
+osm        &     word       &  word        & word     & --      & {\bf 15.82} \\
+osm        &     BPE         &  BPE          & word     & --      & 15.57 \\
\hline
\hline
{\em newstest17}   &&&&&\\
\hline
WMT+back & BPE & BPE & word & -- & {\bf 16.2} \\
opus+osm & BPE & BPE & word & -- & {\bf 17.3} \\
\hline
\end{tabular}
}
\caption{Phrase-based SMT tested on newsdev 2015 and newstest 2017 (lowercased BLEU). Different types of segmentation in source language text (src), target language text (trg) and during minimum-error rate training (tuning): word-based, byte-pair encoding (BPE) and OMorFi-based (morf). Different data sets for training: Europarl and Rapid2016 (WMT), backtranslated Finnish news (back) and all available data sets including parallel corpora from OPUS (opus). Additional component: operation-sequence model (osm).}
\label{tab:SMT-enfi}
\end{table}

The first observation is that BPE (and also supervised morphological
segmentation) is not very helpful. This is somewhat surprising as we expect a
similar problem as with neural MT in the sense that the productive and rich
morphology in Finnish causes problems due to data sparseness. We can see that
some models benefit from BPE (see {\em back} and {\em opus}) especially if
tuning is done on the word level and not on BPE-segmented output. However, we
have to admit that we did not attempt to optimize the segmentation level and
it can well be that the small BPE vocabulary in our setup is not working well
for SMT.

Another observation is that the operation-sequence model does not lead to significant (or any) improvements. This is in contrast to related work and may be due to data sparseness again due to the morphological richness of Finnish.

The biggest surprise is the positive effect of backtranslated data. The models trained on those data sets only are in fact better than the ones trained on the official training data provided by WMT. This demonstrates the strong domain mismatch between training and test data and the use of in-domain data, even very noisy ones, seems to lead to visible benefits. In combination, we can see substantial improvements over the individual models, which demonstrates the use of backtranslation even for SMT.

Another common outcome in SMT is the strong impact of language models. We can confirm this once again. Adding a second language model trained on common-crawl data (CC) has a strong influence on translation quality as we can see by the BLEU scores in Table~\ref{tab:SMT-enfi}.

In the manual evaluation, our best SMT system shared 6th rank with four other
systems (interestingly a mix of phrase-based, rule-based and neural systems),
of which two were constrained like ours.

\subsection{NMT with Pre-translated Data}

We were also interested in the combination of SMT and NMT using the
pre-translation approach proposed by \citet{niehues-EtAl:2016:COLING}. In
their model, SMT-based translations of the source text are simply concatenated
to the input to make it possible for an NMT system to draw information from
other MT models. \citeauthor{niehues-EtAl:2016:COLING} show that the attention
model is capable of learning the connections between the pre-translated part
and the original source language input to jointly influence the generated
target language translations. The approach is straightforward and interesting
because it may improve the faithfulness (or adequacy) of the translation
engine, which can be a problem in neural encoder--decoder models.

One challenge is that training data has to be translated completely to make it possible to learn the final NMT model. One of the problems discussed by \citeauthor{niehues-EtAl:2016:COLING} is the issue of overfitting to the SMT-based translation if the SMT model is trained on the same data set as will be used for learning the NMT model. They propose to weaken the phrase table by removing longer segments and, hence, reducing the capacity of the SMT model to create very generic translation options.

In our setup, we use a different strategy: Instead of using the same data sets for training and translating, we use the backtranslated news data to train a model that can be used to translate the parallel WMT data (Europarl and Rapid2016). With this, we get the same domain-mismatch as during test time with a realistically weak model that avoids over-trusting its capacity when training the NMT model in pre-translated data. Furthermore, we use a WMT-model trained on Europarl and Rapid2016 to translate the backtranslated news data from English back to Finnish again. The latter may be a problem because of the significant noise added due to the double backtranslation but we do not want to discard the important news data completely.

\begin{figure}[t]
\hspace{-.2cm}\includegraphics[width=1.1\columnwidth]{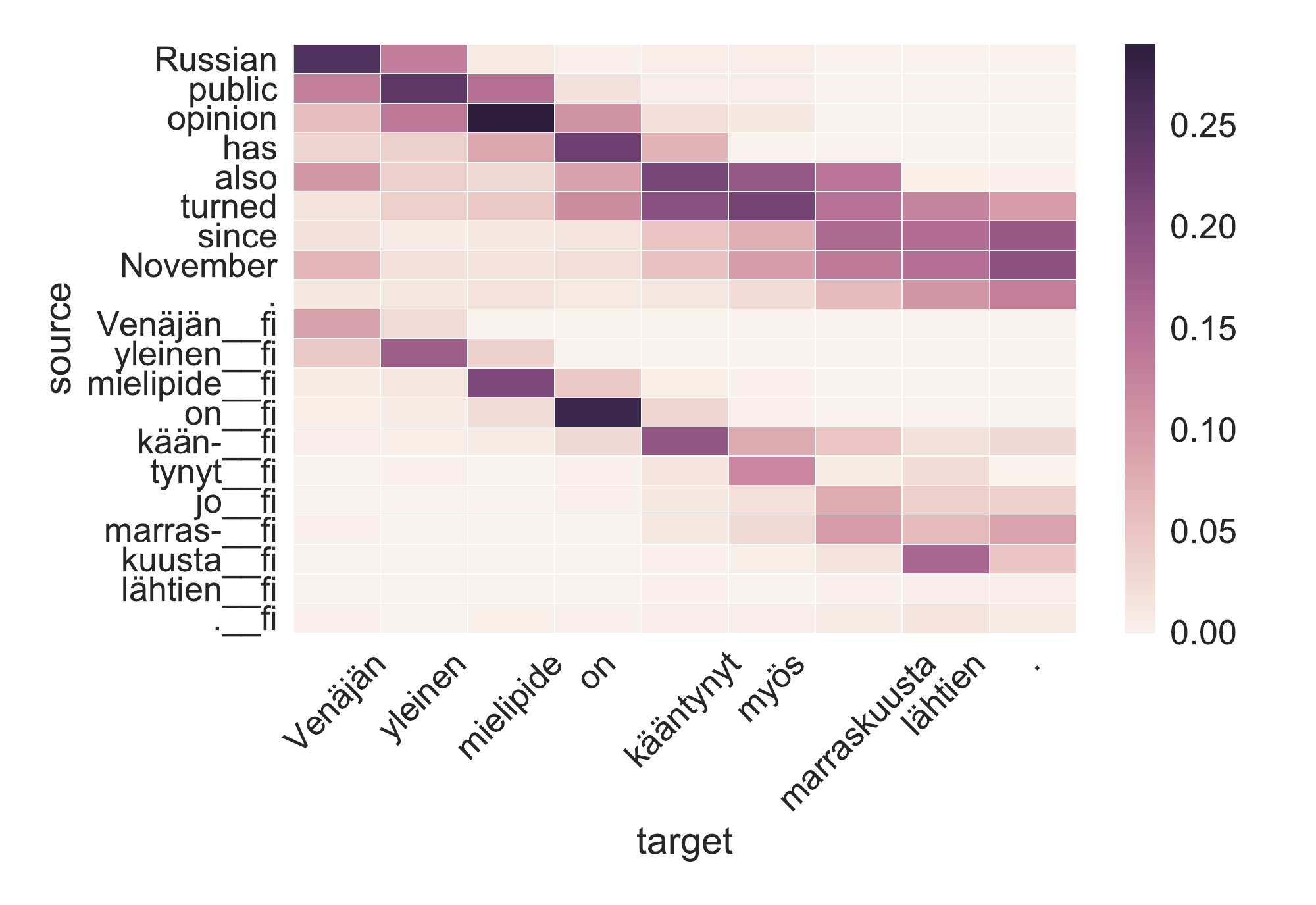}
\caption{Attention with pre-translated data.}
\label{img:premt}
\end{figure}

Another difference in our setup is that we use BPE-segmented SMT models to obtain segmented output that we can use directly to be concatenated with the original (BPE-segmented) source. We mark the pre-translated part with a special suffix and then train a standard attention-based NMT model. We use similar parameters as for our standard NMT experiments: 256-dimensional word embeddings, encoder states and attentions, 512-dimensional decoder states, and a vocabulary of 50,000 in source and target language. It turns out that, indeed, the model learns to look at both the source language and the pre-translated text, as we can see in the attention plot in Figure~\ref{img:premt}.

Unfortunately, the training process is very slow due to the extended input sequence and, hence, converges very slowly. No useful model could be submitted before the official deadline. Our final system tested after the official submission is an ensemble of four independently trained models with savepoint averaging over the last four savepoints and reaches a lowercased BLEU score of 17.34\% on newsdev 2015 and 20.92\% on newstest 2017 in our internal evaluations (but only 19.8\% BLEU in the official on-line system). Even tough this looks quite encouraging compared to the SMT scores, it is still below the plain NMT models, which is, of course, a bit disappointing. However, the results are not directly comparable and there is some variation that needs to be accounted for. More detailed analyses are required to study the possible contributions by the pre-translations. Further investigations of attention plots may reveal whether the model still overfits to the SMT output, which could be a good reason why it underperforms in the end. The additional complexity and the increased length of the input sequences are certainly other reasons for the negative outcome. It also seems that the strong performance of the NMT model also with respect to adequacy make it difficult to improve it further with a weaker SMT model.

\subsection{Manual Evaluation}

The outputs of the best SMT and NMT systems were partially reviewed and
compared by a professional translator. The impression of the reviewer was that
the perceived quality of NMT far exceeds that of SMT, mainly due to the
superior fluency of NMT. The BLEU scores of the systems also indicate a
significant quality difference in favor of NMT. However, single-reference BLEU
scores are known to be unreliable indicators of quality for morphologically
complex languages \cite{Bojar2010TacklingSD}, and they are also known to favor
SMT over other MT methods \cite{Callison-Burch06re-evaluatingthe}. Due to this, it is possible that the BLEU scores,
impressive as they are, do not reflect the real qualitative impact of NMT for
English--Finnish MT.

To explore whether single-reference evaluation underestimates NMT quality, a sample of 68 sentences was extracted from the test set. Both SMT and NMT translations of the sample were postedited with minimal changes to the same quality level as the reference translation. The minimally edited MT was then used as a TER reference to obtain a more reliable estimate of the MT quality. The sample was chosen from sentences where SMT has a sentence-level TER that is at least 10 points lower than the corresponding NMT TER, since such differences can indicate evaluation errors. The sample was also restricted to sentences with an SMT TER lower than 40 to reduce postediting workload and filter out low-quality MT.

When postedited MT was used as a reference, total TER/BLEU for the sample changed from 24.7/50.2 to 12.5/76.0 for SMT and from 48.4/25.0 to 18.3/70.5 for NMT. While the score improved for both SMT and NMT, the improvement is clearly much larger for NMT. The test was then repeated for another sample of 68 sentences from the test set, this time selected from the sentences where NMT had lower sentence-level TER. The purpose of this sample was to see if evaluation errors affect single-reference scores for SMT to the same extent as for NMT. With the second sample, total TER/BLEU changed from 58.9/22.1 to 42.5/39.3 for SMT and from 28.2/48.5 to 12.1/77.01 for NMT, so the result was even more favorable for NMT. While the sample size was small, these results strongly suggest that single-reference BLEU scores indeed underestimate NMT quality.

\section{English--Latvian}

Training models for English--Latvian was a rather spontaneous decision and we
did not spend a lot of time optimizing our settings. Backtranslations were
produced with simplistic Latvian--English models. We used a quickly trained character-level NMT model to translate Latvian news data from 2016 and a standard phrase-based SMT model to translate parts of 2014-2016 news data. The statistics of the backtranslations are given in Table~\ref{tab:backtrans-lven}.

\begin{table}[ht]
\small \centering
\begin{tabular}{|l|rrr|}
\hline
SMT            & Sentences &    Latvian    & English\\
\hline
news2014  & 330,152    & 6,469,914  & 7,611,259\\
news2015  & 330,644    & 6,484,318  & 7,624,202\\
news2016  & 313,180    & 6,161,332  & 7,239,953\\
\hline
NMT           & Sentences &    Latvian    & English\\
\hline
news2016  & 2,059,647 & 33,447,392 & 45,262,908\\
\hline
\end{tabular}
\caption{Backtranslated Latvian news data using SMT and NMT.}
\label{tab:backtrans-lven}
\end{table}

\subsection{NMT Models}

We submitted one NMT system that follows the basic BPE-decoder system for
English--Finnish in terms of model size and training settings. It is trained
on the preprocessed versions of the parallel data and the NMT-based
backtranslations. This system yielded a case-sensitive BLEU score of 16.8. We
again applied hyphen retokenization as a postprocessing step, although it was
less useful here than for Finnish (+0.1 BLEU). Again, we trained four
independent models and used savepoint-averaging. For time reasons and given
the low impact of forward-backward reranking observed for Finnish, we
refrained from submitting such a system for English--Latvian.

\subsection{SMT Baselines}

The SMT models we trained use the provided data sets for training translation
models and language models (including a second language model based on common
crawl data) with the same tools as for our English--Finnish systems. We applied BPE to all data sets again with a rather fine-grained segmentation into 20,000 types on training data. Table~\ref{tab:SMT-enlv} summarizes the results of our models on the \textit{newstest} data from 2017.

\begin{table}[ht]
\centering
{\small
\begin{tabular}{|l|c|}
\hline
{\em newstest 2017}                   & BLEU  \\
\hline
SMT WMT            & 13.29 \\
SMT back           & 11.94  \\
SMT WMT+back  & 13.74 \\
\hline
SMT official score (WMT+back)        & 14.7\\
NMT official score (WMT+back)       & 17.3\\
\hline
\end{tabular}
}
\caption{Statistical MT for English--Latvian tested on newstest 2017
(lowercased BLEU). The {\em official} score in the on-line evaluation system
(lowercased) is surprisingly different from our own evaluations. The manual
evaluation for English--Latvian produced no statistically significant ranking.}
\label{tab:SMT-enlv}
\end{table}

We can see that the backtranslated data sets do not work very well in the Latvian case. A small improvement can be observed when combined with the provided training data but the quality of the backtranslations is too poor to have a strong impact on translation quality.

\section{English--Chinese and Chinese--English}
For English/Chinese, we performed experiments with the HNMT system only. We
trained both English--Chinese and Chinese--English models, using all of the
available parallel training data from the WMT/CWMT news translation task.
After cleaning, 24,954,952 sentence pairs remained.
Using the standard Moses tools, we tokenized and truecased the
English data. Two methods were used for Chinese word segmentation, as detailed
below.

All the models are trained by a hybrid character--word level encoder and a character-level decoder. 
The final submissions are generated by ensembles with parameters averaging. 
The official BLEU scores of these two tasks are shown in Table~\ref{tab:BLEU-enzh}.  
The manual evaluation ranked our system in a shared last place (shared with
four other systems) for
Chinese--English, while it was ranked \#9 (better than two unconstrained online
systems) for English--Chinese.

\subsection{Translating Chinese into English}
Chinese is a language without word boundaries, so word segmentation is
necessary before using our hybrid encoder with Chinese source sentences.
There are
different segmentation methods at different granularities, and they will lead
to different translations. In the work of \citet{Su2017Lattice}, they proposed
a lattice-based recurrent encoder which applied three segmentations
at different granularities (from the CTB, PKU and MSRA corpora).
In our model, we just tried two
segmentations: One is a fine-grained method implemented in Zpar
\cite{Zhang2011SyntacticPU}, the other is a coarse-granularity method by
THULAC \cite{sun2016thulac}. The model with THULAC segmentation achieved a
slightly lower BLEU score compared to the model with Zpar segmentation. Thus,
we did not train more models on THULAC segmentation data after 6-day training.
Unlike our results with English--Finnish translation, our experiments with BPE
using a 30,000 size vocabulary did not yield any improvements.

The final submission uses Zpar for segmentation, a hybrid encoder with
60,000 item vocabulary, and a character-level decoder. We use dimensionalities
of 256 for both word and character embedding, encoder LSTM and attention.  The
decoder uses an LSTM of size 512.
We use a single model with parameters averaged from savepoints at 6, 9, 10, 12 and
14 days to generate the final submission. This is a rather unusual setup and
different from the Finnish and Latvian submissions, but it shows parameter
averaging works even when days have passed between savepoints.
 The beam size in the decoding is set to 10.

\begin{table}
\centering
{\small
\begin{tabular}{|l|c|}
\hline
{\em newstest 2017}                   & BLEU  \\
\hline
English--Chinese            & 23.9 \\
\hline
Chinese--English           & 15.9  \\
\hline
\end{tabular}
}
\caption{HNMT official results on English--Chinese language pair news translation task.}
\label{tab:BLEU-enzh}
\end{table}

\subsection{Translating English into Chinese}
In addition to translating English into Chinese orthography (using Chinese
characters, Hanzi), we also explored translating into romanized Chinese (using
the Pinyin system), and then disambiguating the Pinyin to Hanzi with a 3-gram
language model.  This reduces the vocabulary to the circa 1300 syllables in
Standard Mandarin.  However, the final disambiguation step introduces new
errors that were not outweighed by the easier task of predicting Pinyin
output, and we did not pursue this method.

%

For our official submission, we used a hybrid encoder
with 50,000 vocabulary size, and a character-level decoder. Again, we used a
single model with parameters averaged from savepoints at 6, 7.5 and 11.5 days.


\section{Conclusions}

This paper introduces the Helsinki Neural Machine Translation system (HNMT)
and its succesful application to the news translation task in WMT 2017. The
models we trained handle well the translation into morphologically complex
languages such as Finnish and our submission scored best among the
participants in the English--Finnish task. The evaluations show that the
neural models are superior to the strong SMT baselines that exploit the same
tricks such as backtranslated data and automatic word segmentation. Manual
inspections suggest that the advantage of NMT is even underestimated by
single-reference BLEU scores. We also applied our models to English--Latvian
and English--Chinese (in both directions) with a more moderate success. This is not very surprising for Latvian, for which we only invested about a week to set up the experiments and to train the models. For Chinese, manual evaluation will be important to judge the outcome of our systems fairly.

\section*{Acknowledgments}

We wish to thank the anonymous reviewers, one of whom
provided exceptionally thorough comments. The Finnish IT Center for Science
(CSC) provided computational resources. We would also like to acknowledge the support by NVIDIA and their GPU grant. 
Gongbo Tang is supported by China Scholarship Council (No. 201607110016).

\bibliography{emnlp2017}
\bibliographystyle{emnlp_natbib}

\end{document}